\documentclass[10pt, twocolumn, letterpaper]{article}
\usepackage[top=0.8in,bottom=0.8in,left=0.7in,right=0.7in,columnsep=0.35in]{geometry}
\usepackage{amsmath}
\usepackage{amsthm}
\usepackage[square,sort,comma,numbers]{natbib}
\usepackage{mathrsfs}
\usepackage{graphicx}
\usepackage{color}
\usepackage{xr}

\newtheorem{thm}{Theorem}[section] %(If you want theorem numbered
\newtheorem{lemma}{Lemma}[section] %%    with section number.
\newtheorem{cor}{Corollary}[section]

\newtheorem{definition}{Definition}[section]
\newcommand{\bed}{\begin{definition}}
\newcommand{\eed}{\end{definition}}

\newcommand{\bitem}{\begin{itemize}}
\newcommand{\eitem}{\end{itemize}}

\newcommand{\goto}{\rightarrow}
\newcommand{\margmax}{\mathrm{argmax}}

\newcommand{\beqn}{\begin{equation}}
\newcommand{\eeqn}{\end{equation}}
\newcommand{\balign}{\begin{align}}
\newcommand{\ealign}{\end{align}}

\usepackage{amssymb}
\newcommand{\beq}{\begin{equation}}
\newcommand{\eeq}{\end{equation}}
 
\newcommand{\Hm}{{\cal H}(n,r,k,m,p,q)}

\usepackage[ruled,slide]{algorithm2e}
\newcommand{\Ai}{{\cal A}_{i_1:i_m}} 
\newcommand{\A}{{\cal A}} 
\newcommand{\E}{{\cal E}}

\newcommand{\dY}{d({\cal Y})}
\newcommand{\Yt}{{\cal Y}}
\newcommand{\Yti}{{\cal Y}_{i_1:i_m}}
\newtheorem{remark}{Remark}
\usepackage{hyperref}
\hypersetup{
    colorlinks=true,
    linkcolor=blue,
    filecolor=magenta,      
    urlcolor=cyan,
}
\title{Information Theoretic Limits of Exact Recovery in Sub-hypergraph Models for Community Detection}

\author{Jiajun Liang\\Department of Statistics\\
Purdue University\\West Lafayette, IN 47907, USA\\
\texttt{liang333@purdue.edu}\\
\and
Chuyang Ke\\Department of Computer Science\\
Purdue University\\West Lafayette, IN 47907, USA\\
\texttt{cke@purdue.edu}\\
\and
Jean Honorio\\Department of Computer Science\\
Purdue University\\West Lafayette, IN 47907, USA\\
\texttt{jhonorio@purdue.edu}}

\date{}

\begin{document} 
%%%%%%%%%%%
\maketitle
%%%%%%%%%%%%%
%%%%%%%%%%%%%
%%%%%%%%%%%%%
\begin{abstract}
In this paper, we study the information theoretic bounds for exact recovery in sub-hypergraph models for community detection. We define a general model called the $m-$uniform sub-hypergraph stochastic block model ($m-$ShSBM). Under the $m-$ShSBM, we use Fano's inequality to identify the region of model parameters where any algorithm fails  to exactly recover the planted communities with a large probability. We also identify the region where a Maximum Likelihood Estimation (MLE) algorithm succeeds to exactly recover the    communities with high probability. Our  bounds are tight and pertain to the community detection problems in various models such as the planted hypergraph stochastic block model, the planted densest sub-hypergraph model, and the planted multipartite hypergraph model.
\end{abstract}
\section{Introduction}
Detecting communities has been of great interest in both academia and industry across various fields (e.g., social sciences, genomics, brain connectomics) for decades. A popular approach is to use graph models where clustering algorithms are proposed to recover the membership labels of nodes using the edges' information. Traditional machine learning algorithms for community detection mostly work on ordinary graphs which only model second-order relations. In this case, a co-authorship network\cite{newman2004coauthorship} can only represent pairwise collaboration while it is common to see more than two authors collaborating in one paper. 
To alleviate this issue, higher-order relations are  disassembled into second-order relations and then community detection algorithms are applied on ordinary graphs, thus losing higher-order information. A better approach for modeling higher-order relations is to use a hypergraph where multiple nodes can be connected through a hyperedge. Thus, hypergraph models were proposed to directly study the information of higher-order relations in community detection problems (See e.g.,  \cite{benson2016higher,grilli2017higher}).

A hypergraph ${\cal G}$ consists of a set of $n$ nodes ${\cal V}$ and  a  set of hyperedges $\E$ which is the collection of non-empty subsets  of ${\cal V}$. In the hypergraph ${\cal G}$, we assume that there are $r$ communities each of which has exactly $k$ nodes. The remaining $n-rk$ nodes,  each not in any of the $r$ communities, are called isolated nodes, and could be used to model low-degree nodes of a hypergraph. The $rk$ non-isolated nodes represent the \emph{sub-hypergraph} of major interest. Let $i_1,\cdots,i_m\in {\cal V}$ be a tuple of $m$ distinct nodes. The $m$ nodes $i_1,\cdots,i_m$ are assumed to be connected by a hyperedge, i.e., $(i_1,...i_m) \in {\cal E}$ independently with probability $p_n(m)$ if they are in the same community and with probability $q_n(m)$ otherwise, for all $2\leq m \leq k$. We call such 
a model, a Sub-hypergraph Stochastic Block Model (ShSBM).  We say a hypergraph is $m-$uniform if each hyperedge in $\E$ consists of exactly $m$ distinct nodes and correspondingly, we have a $m-$uniform Sub-hypergraph Stochastic Block Model ($m-$ShSBM) associated with a $m-$uniform hypergraph. When $m=2$ and $n=rk$, the $m-$ShSBM reduces to the classical Stochastic Block Model (SBM). It is of primary interest to consider community detection problem for a  $m-$ShSBM. 

Although there has been some prior research on community detection in uniform hypergraphs, few of the previous works investigate the role of low-degree nodes in the hypergraph (i.e., they assume $n-rk=0$.)  and most importantly the corresponding information theoretic limits. \cite{chien2018community,kim2018stochastic,lin2017fundamental} considered the asymptotic statistical  lower bound instead of the more challenging finite sample information lower bound and did not model isolated nodes. \cite{corinzia2020statistical} modelled isolated nodes and assumed that the hyperedges have Gaussian weights which is mathematically interesting, but not a typical assumption in practice, due to the lack of applications in the real-world for this regime. \cite{ahn2019community,cole2020exact,ahn2018hypergraph} focused on stochastic block models and did not model isolated nodes. The aforementioned works only focus on the stochastic block model in hypergraphs. \cite{ke2020exact} provided an algorithm for community detection for the densest sub-hypergraph model but provided an upper bound  that is not tight.  In practice, the obtained hypergraph data usually contains a large number of small groups consisting of few nodes, which makes the community detection problem harder. Thus, our goal is to investigate the information theoretic limits of  community detection in uniform hypergraphs considering low-degree nodes. 

In this paper, we study the information theoretic bounds in a $m-$ShSBM. Instead of focusing on binary hypergraphs, we assume that each tuple of $m$ distinct nodes is assigned with a random hyperedge weight on $[0,1]$. Intuitively speaking, if the hyperedge weight follows a Bernoulli distribution, this will reduce to the canonical topological setting  in which we model the presence or absence of the hyperedge (i.e., $m$-order relations). If the hyperedge weight follows a continuous distribution, we can model the heterogeneous strength of the connection between nodes (See e.g, \cite{tian2009hypergraph,ahn2018hypergraph}). Hence, in what follows, we assume that each $m-$nodes tuple  is independently  assigned with a random hyperedge value with expectation parameter $p_n(m)$  if the $m$ distinct nodes are in the same community and with expectation parameter $q_n(m)$ otherwise. For simplicity, we fix $n$ and $m$ and replace the notations $p_n(m)$ and   $q_n(m)$ with $p$ and $q$ respectively below whenever it is clear from the context. Intuitively, when the size of each community $k$ and the difference between $p$ and $q$ are small, exact recovery will be hard as it will be difficult to distinguish the sub-hypergraph from isolated nodes. This gives rise to the following two of the most important questions of community detection in a $m-$ShSBM: 
\begin{itemize}
    \item  (i) Under what configurations of the model parameters $n,m,r,k,p,q$, any algorithm will fail in exact recovery of communities in the $m-$ShSBM with a large probability.
    \item  (ii) Under what configurations of the model parameters $n,m,r,k,p,q$, a statistically powerful algorithm using Maximum Likelihood Estimation (MLE) will succeed in exact recovery with high probability.
\end{itemize}
Note that we aim to investigate the statistical limits of the two questions as the algorithm in (ii) may be computationally hard. Finding a powerful and efficient community detection algorithm under our setting is still an open question and we leave it for future work.

 Our contributions are two-fold:  To the best of our knowledge, we provide the first information theoretic limits of exact recovery of communities in the sub-hypergraph model with bounded random weights.  Recall that $p$ and $q$ are the mean parameters of hyperedge weights within and not within the same community, respectively. We show that the minimax lower bound condition for any algorithm is  $\frac{(p-q)^2}{(p\wedge q)(1-p\vee q)}=O(k\log(n/k)/\tbinom{k}{m})$\footnote{Here $p\wedge q\equiv\min\{p,q\}$ and $p\vee q\equiv\max\{p,q\}$. }\label{note1}. We also show that the MLE algorithm upper bound condition is  $\frac{(p-q)^2}{(p\vee q) (1-p\wedge q)}=\Omega(k\log(n)/\tbinom{k}{m})$\textsuperscript{\ref{note1}}. Note that $p\wedge q(1-p\vee q)\leq p\vee q (1-p\wedge q)$ and thus our upper bound and lower bound are tight up to a $\log(k)$ term.
\section{Preliminaries}
In this section, we summarize the  formal notations and definitions used throughout this paper.
\subsection{Notations}
For short, we write  $[n]=\{1,\ldots,n\}$ for any integer $n$, $p\wedge q=\min\{p,q\}$ and $p\vee q=\max\{p,q\}$. We denote  the cardinality of the set $S$ by $|S|$. We say $h(n)=\Omega(g(n))$ or $g(n)=O(h(n))$ if $\lim_{n\goto +\infty}h(n)/g(n)\geq c_0$ for some constant $c_0$, and  $g(n)=o(h(n))$ if $\lim_{n\goto +\infty}h(n)/g(n)\goto +\infty$.  Let $\mathbb{I}_{\{E\}}$ be the indicator function such that $\mathbb{I}_{\{E\}}=1$ iff the event $E$ is true and $\mathbb{I}_{\{E\}}=0$, otherwise. Let ${\cal X}\subseteq [0,1]$ and ${\cal F}$ be the set of probability distributions 
supported on ${\cal X}$. Let ${\cal F}(p)$ be a subset of of ${\cal F}$, corresponding to the set of probability distributions with mean $p$ and sub-Gaussian norm $\sigma_p$ \cite{buldygin1980sub}.
\begin{remark}
Note that the mean and sub-Gaussian norm of any bounded random variable are well-defined and exist.
\end{remark}
\subsection{Definitions}
Recall that a hypergraph ${\cal G}=({\cal V},\E)$ on $n$ nodes consists of a node set ${\cal V}$ and a hyperedge set $\E$, where ${\cal V}=[n]$ and each element of $\E$ is a non-empty subset of ${\cal V}$.  For $m\leq n$, we call a hypergraph ${\cal G}$ \emph{ m-uniform} if $|e|=m$ for all $ e \in \E$. Further, let ${\cal A}$ be a $m-$order $n-$dimensional symmetric tensor. That is for any distinct $ i_1,\cdots,i_m\in {\cal V}$, we define $\A_{i_1\cdots i_m}$ as the hyperedge weight assigned to the $m$ nodes $i_1,\ldots,i_m$ (e.g., in the binary hypergraph, $\A_{i_1\cdots i_m}=1$ if there is a hyperedge connecting nodes $i_1,\ldots, i_m$ and $\A_{i_1\cdots i_m}=0$ otherwise). Therefore, ${\cal A}$ is the hyperedge weights tensor related to the hypergraph ${\cal G}$, an extension of the adjacency matrix of an ordinary graph. Now, we introduce the set $Y$ of membership partition hypotheses $y \in [r+1]^n$ as follows
\begin{equation*}
Y=\bigg\{y\in [r+1]^n:
\begin{array}{ll} 
&y_i\in[r+1],\\
&\sum^{r+1}_{c=1}\mathbb{I}_{\{y_{i}=c\}}=1,\, i \in [n],\\
&\sum^n_{i=1}\mathbb{I}_{\{y_{i}=c\}}=k,\, c \in [r].  \\
\end{array}\bigg\},
\end{equation*}
where  if $y_i=j$, node $i$ belongs to community $j$ for $j\in [r]$, and if $y_i=r+1$, node $i$ is an isolated node in the partition hypothesis $y$. It can be observed that $Y$ is the collection of all possible membership partition hypotheses that divide  $n$ nodes into $n-rk$ isolated nodes and cluster the remaining $rk$ nodes into $r$ communities of equal size $k$. Denote the ground truth partition hypotheses by $y^*=(y^*_1,\ldots, y^*_n)$, where $y^*_i$ is the ground truth label of node $i$. Next, we introduce the formal definition of $m-$ShSBM below.
\bed
\emph{  $m$-uniform  sub-hypergraph Stochastic Block Model ($m-$ShSBM)}.  For a $m-$ShSBM on $n$ nodes, we assume that, there is a ground truth membership partition hypothesis $y^*\in Y$.  Conditioned on $y^*$, we further assume that the weights of the hyperedges are independent random variables on $[0,1]$. For $m$ distinct nodes $i_1,\ldots,i_m \in {\cal V}$, the weight ${\cal A}_{i_1\ldots i_m}$ is sampled as follow
\begin{align*}
{\cal A}_{i_1\ldots i_m}|y^*\sim \bigg\{
\begin{array}{ll} 
f(p), &\qquad \mbox{if $y^*_{i_1}=\ldots=y^*_{i_m}\in [r]$ }, \\
f(q),  &\qquad \mbox{otherwise},  \\
\end{array}
\end{align*}
where $f(p)\in {\cal F}(p)$ and  $f(q)\in {\cal F}(q)$. Here $f(p)$ (or $f(q)$) is a probability distribution with mean $p$ (or $q$). 
\eed
We assume that the mean parameters fulfill $p\neq q$ and denote the above $m-$ShSBM by $\Hm$. 
\begin{remark}
Unless specified otherwise, we allow $r,k,m,p,q$ to vary with $n$. We assume that $r,k$ are known quantities in the algorithms of community detection.
\end{remark}
\begin{remark} \label{Remark:Models}
Our settings cover the following classical models when the hyperedge weights follow Bernoulli distribution:
\end{remark}
\begin{itemize}
    \item {\bf Planted Hypergraph Stochastic Block model:} Let $n=rk$, $0\leq q< p\leq  1$, $r\geq 2$. There are $r$ planted communities of equal size in the $m-$uniform hypergraph, which has been widely studied in the literature (See e.g., \cite{kim2018stochastic,cole2020exact}).
    \item {\bf Planted Densest Sub-hypergraph Model: } Let $0\leq q<p\leq 1$ and $r=1$. There is a densest sub-hypergraph planted in the hypergraph, which generalizes the  sub-graph model in \cite{chen2014statistical} and is somewhat similar to  \cite{corinzia2020statistical} except that they assume Gaussian-distributed weights.
    \item {\bf Planted Multipartite Hypergraph Model: }Let $n=rk$ and $0<p<q<1$. This generalizes the multipartite graph models (e.g.,\cite{abbe2017community,bar2018block})  to hypergraph models.
\end{itemize}

\subsection{Problems}
We introduce the following concepts of impossible exact recovery and possible exact recovery.
Recall that $\A$ is the hyperedge weights tensor of a  hypergraph sampled from $\Hm$. Let $\hat{y}(\cdot)$ denote a membership estimator and  $\hat{y}(\A)$ denote the estimate of the ground truth membership partition hypothesis $y^*$ given the hyperedge weights tensor $\A$. 
\bed
\emph{ Impossible exact recovery}.  We say exact recovery is \emph{ impossible} for the configuration $\Hm$ if it satisfies that for any algorithm $\hat{y}(\cdot)$ that takes ${\cal A}$ as input and outputs an estimate $\hat{y}({\cal A})$, we have $\mathbb{P}(y^*\neq\hat{y}({\cal A})|y^*)=1-o_n(1)$.
\eed
\bed
\emph{ Possible exact recovery}.  We say exact recovery is \emph{ possible} for the configuration $\Hm$ if it satisfies that there exists an algorithm that takes $\A$ as input and outputs an estimate $\hat{y}(\A)$ such that $\mathbb{P}(y^*=\hat{y}(\A)|y^*)=1-o_n(1)$.
\eed

Our goal is to investigate the conditions (i.e.,  configurations of $\Hm$ ) where we could have possible (or impossible) exact recovery of communities.

\section{Main Results}\label{mainresult}
In this section, we summarize our  results for impossible exact recovery and possible exact recovery in a $m-$ShSBM. 
\subsection{Lower Bounds}
We first present our impossibility results for exact recovery. Recall that $Y$ is the set of all possible membership partition hypotheses, $y^*$ is the ground truth hypothesis and $\A$ is the random hyperedge weights tensor sampled from $\Hm$. The following lemma is provided by applying the famous Fano's inequality. 
\begin{lemma}[Theorem 2.10.1  in \cite{Cover06}]\label{mlemma:LB1} Assume that $y^*$ is chosen from $Y$ uniformly at random. 
We have that, 
\begin{equation}
        \mathbb{P}(\hat{y}(\A)\neq y^*)\geq 1-\frac{\mathbb{I}(y^*,\A)+\log(2)}{\log(|Y|)},
\end{equation}
where $\mathbb{I}(y^*,\A)$ is the mutual information between $y^*$ and $\A$.
\end{lemma}

By elementary combinatorial calculations, the cardinality of the hypothesis class $|Y|=\frac{n!}{(n-rk)!(k!)^r}$.  Therefore, to get a lower bound, it is sufficient to calculate the mutual information. Recall that $f(p)$, $f(q)$ are the distributions of hyperedge weights within and not within the same community, respectively. We have the following bound for the mutual information $\mathbb{I}(y^*,\A)$.
\begin{lemma} \label{mlemma:LB2}  Assume that $y^*$ is chosen from $Y$ uniformly at random.
For a hyperedge weights tensor $\A$ sampled from $\Hm$, we have that,
\begin{equation}
     \mathbb{I}(y^*,\A)\leq r d(p,q)\tbinom{k}{m}(1-r\frac{\tbinom{n-m}{k-m}}{\tbinom{n}{k}}),
\end{equation}
where $d(p,q)=\mathbb{KL}(f(p)|f(q))+\mathbb{KL}(f(q)|f(p))$.
\end{lemma}

Here $d(p,q)$ is the symmetric KL-divergence that measures the difference between the weights distribution within the same community and that not within the same community.  

Combining Lemma \ref{mlemma:LB1}-\ref{mlemma:LB2}, we have the following  impossibility result, which identifies a region where any community detection algorithm fails in exact recovery for any distribution of hyperedge weights supported on $[0,1]$, with probability at least $c_0-o_n(1)$.
\begin{thm}[Lower bound with uniform prior]\label{thm:LB}Assume that $y^*$ is chosen from $Y$ uniformly at random.
For a hyperedge weights tensor $\A$ sampled from $\Hm$, if 
\begin{equation} \label{LB:cond1}
d(p,q)\leq (1-c_0)(\frac{k\log(n/k)}{\tbinom{k}{m}}),
\end{equation}
then exact recovery for any algorithm $\hat{y}(\cdot)$ fails with a large probability. More formally, 
\[\inf_{\hat{y}(\cdot)}\mathbb{P}(\hat{y}({\cal A})\neq y^*)\geq c_0-o_n(1).\]
\end{thm}
Based on this theorem, if we take ${\cal F}$ be the class of Bernoulli distributions, we have the following minimax lower bound for the classical binary hypergraph without assuming that the ground truth hypothesis $y^*$ is chosen from $Y$ uniformly.
\begin{thm}[Minimax lower bound for binary hypergraphs]\label{thm:minimaxLB} For an adjacency tensor $\A$ sampled from $\Hm$, if
\begin{equation} \label{LB:cond2}
  \frac{(p-q)^2}{(p\wedge q) (1-p\vee q)}\leq (1-c_0)(\frac{k\log(n/k)}{\tbinom{k}{m}}),
\end{equation}
then 
\begin{equation*}
    \inf_{\hat{y}(\cdot)}\sup_{y^*\in Y}\mathbb{P}(\hat{y}({\cal A})\neq y^*|y^*)\geq c_0-o_n(1).
\end{equation*}
\end{thm}
Directly,  Theorem \ref{thm:minimaxLB} yields the minimax lower bound for the specific binary models discussed in Remark \ref{Remark:Models}: 
\begin{itemize}
    \item {\bf Planted Stochastic Block Model:} $(p-q)^2/(p(1-q))=O(k\log(r)/\tbinom{k}{m})$.
    \item {\bf Planted Densest Sub-hypergraph Model: } $(p-q)^2/(p(1-q))=O(k\log(n/k)/\tbinom{k}{m})$.
    \item {\bf Planted Multipartite Hypergraph Model: } $(p-q)^2/(p(1-q))=O(k\log(r)/\tbinom{k}{m})$.
\end{itemize}
Theorem \ref{thm:minimaxLB}  also provides the minimax lower bound considering a general class of distributions ${\cal F}$.
\begin{cor}[Minimax lower bound for weighted hypergraphs]\label{cor:minimaxLBw} For a hyperedge weights tensor $\A$ sampled from $\Hm$, if 
\begin{equation} \label{LB:cond3}
  \frac{(p-q)^2}{(p\wedge q) (1-p\vee q)}\leq (1-c_0)(\frac{k\log(n/k)}{\tbinom{k}{m}}),
\end{equation}
then 
\begin{equation*}
    \inf_{\hat{y}(\cdot)}\sup_{ {\cal F},y^*\in Y}\mathbb{P}(\hat{y}({\cal A})\neq y^*|y^*)\geq c_0-o_n(1).
\end{equation*}
\end{cor}
This corollary states that if the condition (\ref{LB:cond3}) holds, then for any community detection algorithm, there exists a corresponding hypothesis $y^*$ such that the exact recovery of communities fails with a probability at least $c_0-o_n(1)$. As a result,  exact recovery would be impossible in the minimax sense provided that $(p-q)^2/\big(p\vee q (1-p\wedge q)\big)=o(k\log(n/k)/\tbinom{k}{m})$. 

Note that the condition (\ref{LB:cond3}) will be easier to satisfy with a large value of $n$ and a small difference between weights mean parameters $p$ and $q$. This is in line with our intuition, as the sub-hypergraph of interest would be more difficult to be distinguished from the isolated nodes. Another observation is that the condition is also easier to satisfy given a large value of $m$ when $2\leq m \leq \lfloor \frac{k}{2} \rfloor$ and a small value of $m$ when $\lceil \frac{k}{2} \rceil \leq m \leq k$, which is a more precise result compared to  the corresponding conditions $2\leq m \leq \lfloor \frac{n}{2} \rfloor$ and $\lceil \frac{n}{2} \rceil \leq m \leq n$ in \cite{ahn2018hypergraph}.
%%%%%%%%%%%%%%%%%%%%%%%%%%%
%%%%%%%%%%%%%%%%%%%%%%%%%%%
%%%%%%%%%%%%%%%%%%%%%%%%%%%
\subsection{Upper Bound}
To examine the tightness of the lower bound, we provide an algorithm using Maximum Likelihood Estimation (MLE) for $p>q$. This algorithm is an extension of the MLE algorithm in \cite{chen2014statistical}. For multipartite hypergraph models where $q>p$, we can change the maximization in Algorithm \ref{UBalgorithm1} into minimization. 
\begin{algorithm}\label{UBalgorithm1}
{\bf Input:}The hyperedge weights tensor $\A$\\%{}
{\bf Output:}The estimated membership of all nodes $\hat{y}(\A)$
\begin{align}
\begin{split} \label{UB}
       \hat{y}(\A)=\margmax_{y\in Y}\sum_{\substack{1\leq i_1<\cdots<i_m\leq n}}&{\cal A}_{i_1\ldots i_m}\mathbb{I}_{\{y_{i_1}=\ldots=y_{i_m}\in [r]\}}
\end{split}
\end{align}
 \caption{MLE Solution for $m-$ShSBM ($q<p$)}
\end{algorithm}
Note that  Algorithm \ref{UBalgorithm1} exhaustively explores all the hypotheses, which is computationally NP-hard. We leave the search of a statistically powerful and computationally efficient algorithm on a $m-$ShSBM for future work.

In particular, we stress that Algorithm \ref{UBalgorithm1} works for any hypergraph data where we do not specify that the bounded hyperedge weights follow any specific distribution. The following theorem states the upper bound for exact recovery attained by Algorithm \ref{UBalgorithm1}. 

\begin{thm}[Upper Bound]\label{thm:UB}  For a hyperedge weights tensor $\A$ sampled from $\Hm$, if %$\Hm$ satisfies
$q<p$ and
\begin{equation}\label{UBcond1}
    \frac{(p-q)^2}{p(1-q)}= \Omega(\frac{k\log(n)}{\tbinom{k}{m}}),
\end{equation}
alternatively, if $p<q$ and
\begin{equation}\label{UBcond2}
    \frac{(p-q)^2}{q(1-p)}= \Omega(\frac{k\log(n)}{\tbinom{k}{m}}),
\end{equation}
then exact recovery of communities is possible.
\end{thm}

We conclude from Theorem \ref{thm:UB} that  if the mean parameters of the random hyperedge weights satisfy 

\begin{equation}\label{UBcond}
  \frac{(p-q)^2}{(p\vee q)(1-p\wedge q)}= \Omega(k\log(n)/\tbinom{k}{m}),  
\end{equation}
 then exact recovery is possible. It can be observed that when the difference between $p$ and $q$ and the value of $k$ are both large, this condition will be easier to satisfy, which is in line with our intuition. 
The upper bound results for the specific binary models discussed in Remark \ref{Remark:Models}
directly follow from 
condition (\ref{UBcond}):
\begin{itemize}
    \item {\bf Hypergraph Stochastic Block Model:} $(p-q)^2/(p(1-q))=\Omega(k\log(r)/\tbinom{k}{m})$.
    \item {\bf Densest Sub-hypergraph Model: } $(p-q)^2/(p(1-q))=\Omega(k\log(n)/\tbinom{k}{m})$.
    \item {\bf Multipartite Hypergraph Model: } $(p-q)^2/(q(1-p))=\Omega(k\log(rk)/\tbinom{k}{m})$.
\end{itemize}

Note that one may obtain a tighter upper bound if more information about the distribution of hyperedge weight is available. The following theorem states the upper bound if the sub-Gaussian norms of the hyperedge weights are available.  

\begin{thm}[Upper Bound given sub-Gaussian norm]\label{thm:UB2}  For a hyperedge weights tensor $\A$ sampled from $\Hm$, if 
\begin{equation}\label{UBcond3}
    \frac{(p-q)^2}{\max\{\sigma^2_p,\sigma^2_q\}}= \Omega(\frac{k\log(n)}{\tbinom{k}{m}}),
\end{equation}
where $\sigma^2_p$ and $\sigma^2_q$ are the sub-Gaussian norms of the hyperedge weight distributions within and not within the same community respectively, then exact recovery of communities is possible. 
\end{thm}
In fact, the Bernoulli random variable is the one with the largest variance among all random variables with a prescribed mean on $[0,1]$.  Theorem \ref{thm:UB2} characterizes information theoretic upper bound for exact recovery in a $m-$ShSBM with small-variance hyperedge weights by leveraging the sub-Gaussian norm (proxy of variance).
\subsection{Comparison to the literature}
\cite{ahn2018hypergraph} considered the exact recovery of a $m-$ShSBM where $q<p$ and there are no isolated nodes. 
 They provided an upper bound $(p-q)^2/p=\Omega(r^{m-1}n\log(n)/(m\tbinom{n}{m}))$, while our Theorem \ref{thm:UB} implies $\frac{(p-q)^2}{(p\vee q)(1-p\wedge q)}= \Omega(k\log(n)/\tbinom{k}{m})$, providing a wider region and an extra angle on the role of the size of the sub-hypergraph. \cite{cole2020exact} considered a $m-$ShSBM where there are no isolated nodes, hyperedges are binary and $m$ does not vary with $n$. They assumed that  $q<p$ and  characterized the minimax lower bound  by $\max\{\mathbb{KL}(\mbox{Ber}(p)|\mbox{Ber}(q)),\mathbb{KL}(\mbox{Ber}(q)|\mbox{Ber}(p))\}=O( \log(n-k)/\tbinom{k-1}{m-1})$, while our Theorem \ref{thm:LB}-\ref{thm:minimaxLB} implies $\mathbb{KL}(\mbox{Ber}(p)|\mbox{Ber}(q))+\mathbb{KL}(\mbox{Ber}(q)|\mbox{Ber}(p))= O(k\log(n/k)/\tbinom{k}{m})$, 
  providing a wider region and being more precise in recognizing the role of $m$. That is, the lower bound in \cite{cole2020exact} contains an extra factor $m$. Finally, to the best of our knowledge, our Theorem \ref{thm:UB2} is entirely novel, and thus, there is no possible comparison with prior literature. 
\section{Proofs}
In this section, we only prove Theorem \ref{thm:UB}. The proofs for Lemmas (\ref{mlemma:LB1}-\ref{mlemma:LB2}), Theorems (\ref{thm:LB}-\ref{thm:minimaxLB}) and (\ref{thm:UB2}) and Corollaries \ref{cor:minimaxLBw} are put in the appendix.
\subsection{Proof of Theorem \ref{thm:UB}}
We only prove the case when $q<p$, the proof for $p<q$ is similar and therefore we omit the proof.

Let ${\cal Y}$ be a tensor with the same size as ${\cal A}$ where $\Yt_{i_1\ldots i_m}=\mathbb{I}_{\{y_{i_1}=\ldots=y_{i_m}\in[r]\}}$ for any $i_1,\ldots, i_m\in[n]$. For short, we write  ${\cal B}_{i_1:i_m}={\cal B}_{i_1\ldots i_m}$  
and $\langle {\cal B},{\cal C}\rangle=\sum_{\substack{1\leq i_1<\cdots< i_m\leq n}}{\cal B}_{i_1:i_m}{\cal C}_{i_1:i_m}$ for any two tensors ${\cal B}$ and ${\cal C}$. We use $\Yt^*$ instead of $\Yt$  in these terms by replacing $y$ with $y^*$ whenever it is clear from the context.

By the construction of Algorithm \ref{UBalgorithm1}, only if for all $y\neq y^*,\langle \A,\Yt\rangle<\langle \A,\Yt^*\rangle$,  exact recovery with the MLE algorithm succeeds. Hence  our goal is to show that under our conditions
\begin{equation}\label{UBpf1:1}
    \mathbb{P}(\exists\, y\neq y^*,\langle \A,\Yt\rangle\geq \langle \A,\Yt^*\rangle|y^*)=o_n(1).
\end{equation}
Note that by our definitions and elementary algebra, for any $y\in Y$
\begin{equation}\label{UBpf2:1}
    \langle \A,\Yt\rangle\geq \langle \A,\Yt^*\rangle \iff \underbrace{(I)}_{\mbox{Noise}}+\underbrace{(II)}_{\mbox{Signal}}\geq 0,
\end{equation}
where 
\begin{align*}
(I)=&\langle \A-\mathbb{E}[\A],\Yt-\Yt^*\rangle,\qquad (II)=\langle \mathbb{E}[\A],\Yt-\Yt^*\rangle.
\end{align*}
In the above, we have a "noise" term $(I)$ which is a random quantity to be analyzed by concentration inequalities, and a "signal" term $(II)$ which is an expectation that measures the difference between $\Yt$ and $\Yt^*$ in terms of $\mathbb{E}[\A]$.
Now, we introduce a disagreement function
\begin{equation*}
    d(\Yt)=\sum_{\substack{1\leq i_1<\cdots<i_m\leq n}}\mathbb{I}_{\{\Yti^*\neq \Yti\}}/2,
\end{equation*}
which counts all the $m-$node tuples $\{i_1,\ldots,i_m\}$ such that $\Yti^*=1$ but $\Yti=0$ (or by symmetry $\Yti^*=0$ but $\Yti=1$).

By our definitions and direct calculations,
\begin{align*}
    (II)=&\sum_{\substack{1\leq i_1<\cdots<i_m\leq n}}\big(q\cdot 1 \cdot \mathbb{I}_{\{\Yti=1,\Yti^*=0\}}\cdot 1\\
    &\qquad\qquad\quad+p\cdot(-1)\cdot\mathbb{I}_{\{\Yti=0,\Yti^*=1\}}\big)\\
    =&-(p-q)d(\Yt).
\end{align*}
It can be observed that (\ref{UBpf1:1}) is equivalent to
\begin{equation}\label{UBpf1:2}
    \mathbb{P}(\cup_{t}E_t|y^*)=o_n(1),
\end{equation} 
where
\begin{equation*}
    E_t=\big\{\exists\, y\in \{y\in Y: d(\Yt)=t\},(I)\geq (p-q)\dY\big\}.
\end{equation*}
Note that $\tbinom{k-1}{m-1}\leq \dY \leq  r\tbinom{k}{m}$ where the first inequality is achieved by swapping one non-isolated node with an isolated node and the latter inequality holds because the number of $m-$node tuples such that $\Yti^*=1$ is bounded by $r\tbinom{k}{m}$.

Therefore, by the union bound $\mathbb{P}(\cup_i E_i|y^*)\leq \sum_i\mathbb{P}(E_i|y^*)$, to show (\ref{UBpf1:2}), it is sufficient to show that
\begin{equation}\label{UBpf1:3}
    \sum_{ \tbinom{k-1}{m-1}\leq t \leq  r\tbinom{k}{m}}\mathbb{P}(E_t|y^*)=o_n(1).
\end{equation}
Fixing $t$, we are going to derive a bound for $\mathbb{P}(E_t|y^*)$. 

To evaluate the probability of $(I)\geq (p-q)\dY$ given hypothesis $y$, we have the following lemma which is proved in the appendix.
\begin{lemma}\label{lemma:UB0}Under the conditions of Theorem \ref{thm:UB}, we have that
\[
\mathbb{P}((I)\geq  (p-q)\dY|y^*,y) \leq 2e^{-\frac{3(p-q)^2}{28p(1-q)}\dY}.
\]
\end{lemma}
Now let $D_t=|\{y\in Y: d(\Yt)=t\}|$ be the size of the set $\{y\in Y: d(\Yt)=t\}$. By the union bound and Lemma \ref{lemma:UB0}, we have that
\begin{equation}\label{UBpf1:4}
    \mathbb{P}(E_t|y^*)\leq D_t\cdot 2e^{-\frac{3(p-q)^2}{28p(1-q)}t}.
\end{equation}
To evaluate the value of $D_t$, we have the following lemma which is proved in the appendix. 

\begin{lemma} \label{lemma:UB1} Under the conditions of Theorem \ref{thm:UB}, we have that
\begin{equation*} 
   D_t\leq n^{8kt/\tbinom{k}{m}}.
\end{equation*}
\end{lemma}
Now we are ready to show (\ref{UBpf1:3}). Combining (\ref{UBpf1:3}-\ref{UBpf1:4}) with Lemma \ref{lemma:UB1} and our condition $  \frac{(p-q)^2}{p(1-q)}\geq Ck\log(n)/\tbinom{k}{m}$, where $C$ is a constant, we have 
\begin{align*}
    \sum_{ \tbinom{k-1}{m-1}\leq t \leq  r\tbinom{k}{m}}\mathbb{P}(E_t|y^*)\leq &   \sum_{ \tbinom{k-1}{m-1}\leq t \leq  r\tbinom{k}{m}} D_t\cdot 2e^{-\frac{3(p-q)^2}{28p(1-q)}t}\\
    \leq &
    \sum_{ \tbinom{k-1}{m-1}\leq t \leq  r\tbinom{k}{m}} n^{\frac{8kt}{\tbinom{k}{m}}} 2e^{-C\frac{k\log(n)}{\tbinom{k}{m}}t}\\
    \leq  &\sum_{ \tbinom{k-1}{m-1}\leq t \leq  r\tbinom{k}{m}}2n^{-Ckt/\tbinom{k}{m}}\\
    \leq &Cn^{-Ck\tbinom{k-1}{m-1}/\tbinom{k}{m}},
\end{align*}
where we note that the RHS is $\leq Cn^{-Cm}=o_n(1)$, which finishes the proof.

\section{Future work}
Based on the interesting findings of this work, we hope to generalize the results of $m-$ShSBM to other settings. We  could consider a general community structure 
where there are more than two mean parameters to characterize the distributions of binary hyperedges across different communities. We could also introduce degree heterogeneity 
by adding a degree parameter to each node such that the distributions of hyperedges within the same community are not necessary identical. We hope to find a computationally efficient community detection algorithm for $m-$ShSBM. Although, while heuristics are easy to be developed, coming up with an algorithm with formal guarantees is challenging, and thus, we leave this for future work. 

\bibliographystyle{plain}
\bibliography{main}

\clearpage
\appendix
\onecolumn
\section{Detailed Proofs}
In the appendix, we only prove  Lemmas (\ref{mlemma:LB1}-\ref{lemma:UB1}), Theorems (\ref{thm:LB}-\ref{thm:minimaxLB}) and (\ref{thm:UB2}) and Corollaries \ref{cor:minimaxLBw}. Note that Theorem \ref{thm:UB} is proved in the main file.
\subsection{Proof of Lemma \ref{mlemma:LB1} }
Recall that $Y$ is the set of all possible membership partition hypotheses. Denote $ \mathbb{P}(\hat{y}(\A)\neq y^*)$ by $P_e$. By Fano's inequality \cite[Theorem 2.10.1]{Cover06}, we have 
\begin{equation*}
    \mathbb{H}(P_e)+P_e\log(|Y|)\geq \mathbb{H}(y^*|{\cal A}),
\end{equation*}
where $\mathbb{H}(\cdot)$ is the entropy function. Note that for any binary distribution $F$, $\mathbb{H}(F)\leq \log(2)$. It follows that
\begin{equation*}
    P_e\geq \frac{\mathbb{H}(y^*|{\cal A})-\log(2)}{\log(|Y|)},
\end{equation*}
By  property of the mutual information, we have $\mathbb{H}(y^*|{\cal A})=\mathbb{H}(y^*)-\mathbb{I}(y^*,{\cal A})$. At the same time, by our assumption that $y^*$ is uniformly chosen from $Y$, then by direct calculations $\mathbb{H}(y^*)=\log(|Y|)$ and therefore 
\begin{equation*}
    P_e\geq 1-\frac{\mathbb{I}(y^*,{\cal A})+\log(2)}{\log(|Y|)},
\end{equation*}
which finishes the proof.
\subsection{Proof of Lemma \ref{mlemma:LB2}}
Recall that  $y^*$ is chosen uniformly at random from $Y$. By the convexity of $\mathbb{KL}$ divergence and the definition of  mutual information (See \cite[Page 428 ]{yu1997assouad}), we have
\begin{equation} \label{convex}
     \mathbb{I}(y^*,\A)\leq\frac{1}{|Y|^2}\sum_{y\in Y}\sum_{y'\in Y}\mathbb{KL}(\mathbb{P}_{\A|y}|\mathbb{P}_{\A|y'}),
\end{equation}
where $\mathbb{P}_{\A|y}$ denotes the distribution of the hyperedge weights tensor ${\cal A}$ given membership partition hypothesis $y$. Note that the values of the hyperedges are conditionally independent given the membership partition hypothesis $y$ ( or $y'$). Let $\widetilde{\E}=\big\{e:e\subset {\cal V};\, |e|=m\big\}$ be the set of all possible hyperedges, it follows that,
\begin{equation*}
    \mathbb{KL}(\mathbb{P}_{\A|y}|\mathbb{P}_{\A|y'})=\sum_{e\in \widetilde{\E}}\mathbb{KL}(\mathbb{P}_{\A_e|y}|\mathbb{P}_{\A_e|y'}),
\end{equation*}
and
\begin{equation*}
    \mathbb{I}(y^*,\A)\leq \frac{1}{|Y|^2}\sum_{y\in Y}\sum_{y'\in Y}\sum_{e\in \widetilde{\E}}\mathbb{KL}(\mathbb{P}_{\A_e|y}|\mathbb{P}_{\A_e|y'}).
\end{equation*}
Recall that $f(p)\in {\cal F}(p)$ and $f(q)\in {\cal F}(q)$ are the hyperedge weight distributions within and not within the same community respectively. We can rewrite 
\begin{equation*}
     \mathbb{KL}(\mathbb{P}_{\A_e|y}|\mathbb{P}_{\A_e|y'})=\mathbb{KL}(\mathbb{P}_{\A_e|y}|\mathbb{P}_{\A_e|y'})\sum_{F_1,F_2\in\{f(p),f(q)\}}\mathbb{I}_{\{\A_e|y\sim F_1,\A_e|y'\sim F_2\}}.
\end{equation*}
Note that $\mathbb{KL}(\mathbb{P}_{\A_e|y}|\mathbb{P}_{\A_e|y'})\mathbb{I}_{\{\A_e|y\sim f(p),\A_e|y'\sim f(p)\}}=0$ and that $\mathbb{KL}(\mathbb{P}_{\A_e|y}|\mathbb{P}_{\A_e|y'})\mathbb{I}_{\{\A_e|y\sim f(q),\A_e|y'\sim f(q)\}} =0$, we have
\begin{align*}
         \mathbb{KL}(\mathbb{P}_{\A_e|y}|\mathbb{P}_{\A_e|y'})=&\mathbb{KL}(\mathbb{P}_{\A_e|y}|\mathbb{P}_{\A_e|y'})(\mathbb{I}_{\{\A_e|y\sim f(p),\A_e|y'\sim f(q)\}}+ \mathbb{I}_{\{\A_e|y\sim f(q),\A_e|y'\sim f(p)\}}),
\end{align*}
and thus
\begin{align*}
       \mathbb{I}(y^*,\A)\leq& \frac{1}{|Y|^2}\sum_{y\in Y}\sum_{y'\in Y}\sum_{e\in \widetilde{\E}}\mathbb{KL}(\mathbb{P}_{\A_e|y}|\mathbb{P}_{\A_e|y'})(\mathbb{I}_{\{\A_e|y\sim f(p),\A_e|y'\sim f(q)\}}+\mathbb{I}_{\{\A_e|y\sim f(q),\A_e|y'\sim f(p)\}}).
\end{align*}
By symmetry of $y$ and $y'$, we have
\begin{align*}
       \mathbb{I}(y^*,\A)\leq& \frac{1}{|Y|^2}\sum_{y\in Y}\sum_{y'\in Y}\sum_{e\in \widetilde{\E}}(\mathbb{KL}\big(\mathbb{P}_{\A_e|y}|\mathbb{P}_{\A_e|y'})+\mathbb{KL}(\mathbb{P}_{\A_e|y'}|\mathbb{P}_{\A_e|y})\big)\mathbb{I}_{\{\A_e|y\sim f(p),\A_e|y'\sim f(q)\}}.
\end{align*}
We define $d(p,q)=\mathbb{KL}\big(f(p)|{f(q)})+\mathbb{KL}(f(p)|f(q))$. It follows that
\begin{equation*}
    \mathbb{I}(y^*,\A)\leq \frac{d(p,q)}{|Y|^2}\sum_{y\in Y}\sum_{y'\in Y}\sum_{e\in \widetilde{\E}}\mathbb{I}_{\{\A_e|y\sim f(p),\A_e|y'\sim f(q)\}}.
\end{equation*}

Note that $\sum_{y'\in Y}\sum_{e\in \widetilde{\E}}\mathbb{I}_{\{\A_e|y\sim f(p),\A_e|y'\sim f(q)\}}$ has the same value for any $y\in Y$. The above inequality is then equivalent to 
\begin{align}
\begin{split}\label{LB:1}
           \mathbb{I}(y^*,\A)&\leq \frac{d(p,q)}{|Y|}\sum_{y'\in Y}\sum_{e\in \widetilde{\E}}\mathbb{I}_{\{\A_e|y\sim f(p),\A_e|y'\sim f(q)\}}, \quad \forall\, y \in Y.
\end{split}
\end{align}
Fix $y$. Before continuing, we denote the set of the nodes in  community $i$ of a hypothesis $y$ by $C_i(y)$ and correspondingly the set of all possible hyperedges over the nodes only in the community $i$ of a hypothesis $y$ by $\widetilde{\E}_i(y)$, i.e., $\widetilde{\E}_i(y)=\big\{e\in \E:e\subseteq  C_i(y);\, |e|=m\big\}$. Note that $\A_e|y\sim f(p)$  if and only if $e$ belongs to one of $\widetilde{\E}_1(y),\ldots,\widetilde{\E}_r(y)$. We have
\begin{align*}
    \sum_{e\in \widetilde{\E}}\mathbb{I}_{\{\A_e|y\sim f(p),\A_e|y'\sim f(q)\}}=& \sum_{e\in \widetilde{\E}}\bigg(\mathbb{I}_{\{\A_e|y'\sim f(q)\}}\sum_{i=1}^r\mathbb{I}_{\{e\in \widetilde{\E}_i(y)\}}\bigg)
    = \sum_{i=1}^r\sum_{e\in \widetilde{\E}_i(y)}\mathbb{I}_{\{\A_e|y'\sim f(q)\}}. 
\end{align*}
Note that for all $e\in \widetilde{\E}$, $\sum_{y'\in Y}\mathbb{I}_{\{\A_e|y'\sim f(q)\}}$ has the same value by the symmetry of communities. It follows that for any $e\in \widetilde{\E}_1(y)$
\begin{equation*}
   \sum_{y'\in Y}\sum_{i=1}^r\sum_{e\in \widetilde{\E}_i(y)}\mathbb{I}_{\{\A_e|y'\sim f(q)\}}=r|\widetilde{\E}_1(y)|\sum_{y'\in Y}\mathbb{I}_{\{\A_e|y'\sim f(q)\}}
\end{equation*}
and therefore (\ref{LB:1}) becomes
\begin{equation} \label{ineq1}
    \mathbb{I}(y^*,\A)\leq\frac{rd(p,q)}{|Y|}|\widetilde{\E}_1(y)|\sum_{y'\in Y}\mathbb{I}_{\{\A_e|y'\sim f(q)\}}.
\end{equation}
Fix any $e\in \widetilde{\E}_1(y)$. Note that $\mathbb{I}_{\{\A_e|y'\sim f(q)\}}=1-\mathbb{I}_{\{\A_e|y'\sim f(p)\}}=1-\sum_{i=1}^r\mathbb{I}_{\{e\in \widetilde{\E}_i(y')\}}$. We have
\begin{equation}
    \sum_{y'\in Y}\mathbb{I}_{\{\A_e|y'\sim f(q)\}}=|Y|-\sum_{y'\in Y}\sum_{i=1}^r\mathbb{I}_{\{e\in \widetilde{\E}_i(y')\}}.
\end{equation}
By elementary combinatorial calculations
\begin{align}
\begin{split}\label{LB:8}
        \sum_{y'\in Y}\sum_{i=1}^r\mathbb{I}_{\{e\in \widetilde{\E}_i(y')\}}=&\sum_{i=1}^r\sum_{y'\in Y}\mathbb{I}_{\{e\in \widetilde{\E}_i(y')\}}\\
        =&\sum_{i=1}^r\tbinom{n-m}{k-m}\frac{(n-k)!}{(k!)^{r-1}(n-rk)!}\\
        =&r\frac{\tbinom{n-m}{k-m}}{\tbinom{n}{k}}|Y|
\end{split}
\end{align}
Combining (\ref{ineq1})-(\ref{LB:8}) and noting that $|\widetilde{\E}_1(y)|=\tbinom{k}{m}$ gives
\begin{equation*}
     \mathbb{I}(y^*,\A)\leq r d(p,q)\tbinom{k}{m}(1-r\frac{\tbinom{n-m}{k-m}}{\tbinom{n}{k}}).
\end{equation*}
This finishes the proof.

\subsection{Proof of Theorem \ref{thm:LB}}
Recall that by Lemma \ref{mlemma:LB1}, we have
\begin{equation}\label{LBpf1:1}
        \mathbb{P}(\hat{y}(\A)\neq y^*)\geq 1-\frac{\mathbb{I}(y^*,\A)+\log(2)}{\log(|Y|)},
\end{equation}
where $|Y|$ denotes the size of the set $Y$ and $\mathbb{I}(y^*,\A)$ is the mutual information between $y^*$ and the hyperedge weights tensor $\A$. The goal is to show that under the conditions of this theorem, 
\begin{equation}\label{LBpf1:3}
    \frac{\mathbb{I}(y^*,\A)+\log(2)}{\log(|Y|)}\leq 1-c_0+o_n(1).
\end{equation}

On one hand, by our definitions and elementary combinatorial calculations, 
\begin{equation*}
   |Y|=\frac{n!}{(n-rk)!(k!)^r}. 
\end{equation*}
Further by the inequality $\sqrt{n_0}(\frac{n_0}{e})^{n_0}\leq n_0!\leq  e\sqrt{n_0}(\frac{n_0}{e})^{n_0}$ for any integer $n_0$ and direct calculations,
\begin{equation}\label{LBpf1:2}
    \log(|Y|)\geq  \log(\frac{\sqrt{n}(\frac{n}{e})^{n}}{e\sqrt{ (n-rk)}(\frac{n-rk}{e})^{n-rk}(e\sqrt{ k}(\frac{k}{e})^{k})^r})=\big(  (n-rk)\log(\frac{n}{n-rk})+rk\log(\frac{n}{k})\big)\big(1+o_n(1)\big).
\end{equation}
On the other hand, by Lemma \ref{mlemma:LB2}
\begin{equation}\label{LBpf1:4}
     \mathbb{I}(y^*,\A)\leq r d(p,q)\tbinom{k}{m}(1-r\frac{\tbinom{n-m}{k-m}}{\tbinom{n}{k}}),
\end{equation}
Now inserting (\ref{LBpf1:2}) and (\ref{LBpf1:4})  into the LHS of (\ref{LBpf1:3}) gives
\begin{equation}\label{LBpf1:5}
    \frac{\mathbb{I}(y^*,\A)+\log(2)}{\log(|Y|)}\leq \bigg(\frac{rd(p,q)\tbinom{k}{m}}{rk\log(n/k)}+\frac{\log(2)}{rk\log(n/k)}\bigg)\big(1+o_n(1)\big).
\end{equation}
Note that $rk\log(n/k)\geq \min\{r\sqrt{n}\log(n/k),rk\log(n/\sqrt{n})\}$, we have  
\begin{equation}\label{LBpf1:6}
    \frac{\log(2)}{rk\log(n/k)}\leq \max\{ \frac{\log(2)}{r\sqrt{n}\log(n/k)},\frac{\log(2)}{rk\log(n/\sqrt{n})}\}=o_n(1).
\end{equation}
Combining (\ref{LBpf1:5}-\ref{LBpf1:6})
 with our condition $d(p,q)\leq (1-c_0)(\frac{k\log(n/k)}{\tbinom{k}{m}})$ gives (\ref{LBpf1:3}). This finishes the proof.

 %%%%%%%%%%%%%%%%%
 %%%%%%%%%%%%%%%%%
 %%%%%%%%%%%%%%%%%

\subsection{Proof of Theorem \ref{thm:minimaxLB}}

 We first show that when the hyperedge weights follow Bernoulli distribution, the condition $\frac{(p-q)^2}{(p\wedge q) (1-p\vee q)}\leq (1-c_0)(k\log(n/k)/\tbinom{k}{m})$ implies $d(p,q)\leq (1-c_0)(k\log(n/k)/\tbinom{k}{m})$. 
 
 By our definitions and direct calculations,
 \begin{equation*}
      d(p,q)=(p-q)\log(\frac{p(1-q)}{q(1-p)}).
 \end{equation*}
 
By the inequality $\log(x)\leq x-1$ for all $x>0$ and definitions,  
\begin{equation}\label{minmaxLBpf:0}
    (p-q)\log(\frac{p(1-q)}{q(1-p)})\leq \bigg\{\begin{array}{lr}
        &(p-q)(\frac{p(1-q)}{q(1-p)}-1)=\frac{(p-q)^2}{q(1-p)}, \qquad \mbox{if $p\geq q$,} \\
         &(q-p)(\frac{q(1-p)}{p(1-q)}-1)=\frac{(p-q)^2}{p(1-q)}, \qquad \mbox{if $p\leq q$.}
    \end{array}
\end{equation}
It follows that  
\begin{equation}\label{minmaxLBpf:1}
    d(p,q)\leq \frac{(p-q)^2}{p\vee q(1-p\wedge q)}\leq (1-c_0)(k\log(n/k)/\tbinom{k}{m})
\end{equation}

Now, we are ready to prove the theorem. Assume that $y$ follows distribution $F$ on $Y$ and we do not assume any prior distribution on the ground truth hypothesis $y^*$. By elementary probability
\begin{align*}
    \inf_{\hat{y}(\cdot)}\sup_{y^*\in Y}\mathbb{P}(\hat{y}({\cal A})\neq y^*|y^*)=&\inf_{\hat{y}(\cdot)}\mathbb{E}_{y\sim F}\big[\sup_{y^*\in Y}\mathbb{P}(\hat{y}({\cal A})\neq y^*|y^*)\big]\\
    \geq &  \inf_{\hat{y}(\cdot)}\mathbb{E}_{y\sim F}\big[\mathbb{P}(\hat{y}({\cal A})\neq y|y)\big]\\
    =& \inf_{\hat{y}(\cdot)}\mathbb{P}(\hat{y}({\cal A})\neq y),
\end{align*}
which holds for any distribution $F$ on $Y$. 

Let $F$ be the uniform distribution on $Y$. With condition (\ref{minmaxLBpf:1}) being satisfied, by Theorem \ref{thm:LB}, we have $\inf_{\hat{y}(\cdot)}\mathbb{P}(\hat{y}({\cal A})\neq y^*)\geq c_0-o_n(1)$. It follows that
\begin{equation*}
    \inf_{\hat{y}(\cdot)}\sup_{y^*\in Y}\mathbb{P}(\hat{y}({\cal A})\neq y^*|y^*)\geq c_0-o_n(1).
\end{equation*}
This finishes the proof.
 \subsection{Proof of Corollary \ref{cor:minimaxLBw}}
 Note that
 \begin{equation*}
     \sup_{{\cal F},y^*\in Y}\mathbb{P}(\hat{y}({\cal A})\neq y^*|y^*)\geq \sup_{y^*\in Y}\mathbb{P}(\hat{y}({\cal A})\neq y^*|y^*).
 \end{equation*}
 Taking infimum on both sides with respect to $\hat{y}$, gives
  \begin{equation*}
     \inf_{\hat{y}(\cdot)}\sup_{{\cal F},y^*\in Y}\mathbb{P}(\hat{y}({\cal A})\neq y^*|y^*)\geq \inf_{\hat{y}(\cdot)}\sup_{y^*\in Y}\mathbb{P}(\hat{y}({\cal A})\neq y^*|y^*),
 \end{equation*}
 where we note that by Theorem \ref{thm:minimaxLB}, $\inf_{\hat{y}(\cdot)}\sup_{y^*\in Y}\mathbb{P}(\hat{y}({\cal A})\neq y^*|y^*)\geq c_0-o_n(1)$. This finishes the proof.

%%%%%%%%%%%%%%%%%%%%%
%%%%%%%%%%%%%%%%%%%%%
%%%%%%%%%%%%%%%%%%%%%

%%%%%%%%%%%%%%%%%%%%%%%%
%%%%%%%%%%%%%%%%%%%%%%%%
%%%%%%%%%%%%%%%%%%%%%%%%

\subsection{Proof of Lemma \ref{lemma:UB0}}

Recall that
\begin{equation*}
    (I)=\langle \A-\mathbb{E}[\A],\Yt-\Yt^* \rangle=\sum_{\substack{1\leq i_1<\cdots<i_m\leq n}}(\Ai-\mathbb{E}[\Ai])(\Yti-\Yti^*).
\end{equation*}
By our definitions, we can write it into two parts
\begin{equation*}
    (I)=(Ia)+(Ib),
\end{equation*}
where 
\begin{align*}
    (Ia)=&\sum_{1\leq i_1<\cdots<i_m\leq n}(\Ai-q)\cdot 1 \cdot\mathbb{I}_{\{\Yti=1,\Yti^*=0\}},\\
    \qquad(Ib)=&\sum_{1\leq i_1<\cdots<i_m\leq n}(\Ai-p)\cdot (-1) \cdot\mathbb{I}_{\{\Yti=0,\Yti^*=1\}}.
\end{align*}
It can be observed that the RHS of both $(Ia)$ and $(Ib)$ are a sum of $d(\Yt)$ independent bounded random variables each with mean zero. 

Note that by Bhatia–Davis inequality, the variance of any random variable on $[a,b]$ with mean $\mu$ is bounded by $(\mu-a)(b-\mu)$. It follows that
\begin{equation*}
    \mbox{Var}(\Ai-q|\Yti=1,\Yti^*=0)\leq q(1-q), \qquad\mbox{Var}(\Ai-p|\Yti=0,\Yti^*=q)\leq p(1-p).
\end{equation*}
Hence, by Bernstein's inequality,  for any $t>0$,
\begin{align*}
\mathbb{P}((Ia)\geq t|y^*,y)&\leq e^{-\frac{t^2/2}{\dY q(1-q) +t/3}},\\
\mathbb{P}((Ib)\geq t|y^*,y)&\leq e^{-\frac{t^2/2}{\dY p(1-p) +t/3}}.
\end{align*}
Let $t=\frac{(p-q)}{2}\dY$ and note that $24q(1-q)+4(p-q)\leq 28 p(1-q)$. Therefore,
\begin{align}\label{UBpf2:2}
\mathbb{P}((Ia)\geq \frac{(p-q)d({\cal Y})}{2}|y^*,y)&\leq e^{-\frac{3(p-q)^2}{24q(1-q)+4(p-q)}\dY}\leq e^{-\frac{3(p-q)^2}{28p(1-q)}\dY}.
\end{align}
Similarly for $(Ib)$, we have
\begin{align}\label{UBpf2:3}
\mathbb{P}((Ib)\geq \frac{(p-q)d({\cal Y})}{2}|y^*,y)&\leq e^{-\frac{3(p-q)^2}{24p(1-p)+4(p-q)}\dY}\leq e^{-\frac{3(p-q)^2}{28p(1-q)}\dY}.
\end{align}
Combining (\ref{UBpf2:2}) and (\ref{UBpf2:3}), by the union bound,
\begin{align*}
    \mathbb{P}((Ia)+(Ib)\geq (p-q)d({\cal Y}) |y^*,y)
    \leq &\mathbb{P}((Ia)\geq \frac{(p-q)d({\cal Y})}{2}|y^*,\dY)+\mathbb{P}((Ib)\geq \frac{(p-q)d({\cal Y})}{2}|y^*,\dY)\\
    =&2e^{-\frac{3(p-q)^2}{28p(1-q)}\dY}.
\end{align*}
which holds for any $y\in Y$. Directly, we have
\begin{equation*}
    \mathbb{P}((I)\geq (p-q)d({\cal Y}) |y^*,y)
    \leq 2e^{-\frac{3(p-q)^2}{28p(1-q)}\dY}.
\end{equation*}
This finishes the proof.
%%%%%%%%%%%%%%%%%%%%%%%%
%%%%%%%%%%%%%%%%%%%%%%%%
%%%%%%%%%%%%%%%%%%%%%%%%

\subsection{Proof of Lemma \ref{lemma:UB1}}
To prove the claim of the lemma, we first derive 
a property of $\dY$.

Recall that $y$ ($y^*$) is a (ground truth) membership partition hypothesis, $\Yt_{i_1:i_m}=\mathbb{I}_{\{y_{i_1}=\ldots=y_{i_m}\in[r]\}}$ $\big(\Yt^*_{i_1:i_m}=\mathbb{I}_{\{y^*_{i_1}=\ldots=y^*_{i_m}\in[r]\}}\big)$ and $\dY$ counts all the $m-$node tuples $\{i_1,\ldots,i_m\}$ such that $\Yti^*=1$ but $\Yti=0$. 

Note that $\Yt_{i_1:i_2}=0$ implies $\Yt_{i_1:i_m}=0$.  By our definitions
\begin{align*}
    \dY=&\sum_{\substack{1\leq i_1<\cdots<i_m\leq n}}\mathbb{I}_{\{\Yti=0,\Yti^*=1\}}\\
    \geq & \sum_{\substack{1\leq i_1<\cdots<i_m\leq n}}\mathbb{I}_{\{\Yt_{i_1:i_2}=0,\Yti^*=1\}}\\
    =& \frac{1}{m!}\sum_{\substack{1\leq i_1,\cdots,i_m\leq n\\i_1,\ldots,i_m(distinct)}}\mathbb{I}_{\{\Yt_{i_1:i_2}=0,\Yti^*=1\}}.
\end{align*}
Note that $\Yti^*=1$ implies that nodes $i_1,\ldots,i_m$ are in the same community. For each $(i_1,i_2)$, there are $(m-2)!\tbinom{k-2}{m-2}$ choices of $(i_3,\ldots,i_m)$. It follows that
\begin{equation*}
    \dY\geq \frac{(m-2)!\tbinom{k-2}{m-2}}{m!}\sum_{\substack{1\leq i_1,i_2\leq n\\i_1\neq i_2}}\mathbb{I}_{\{\Yt_{i_1:i_2}=0,\Yt^*_{i_1:i_2}=1\}}=\frac{\tbinom{k}{m}}{\tbinom{k}{2}}\sum_{\substack{1\leq i_1<i_2\leq n\\}}\mathbb{I}_{\{\Yt_{i_1:i_2}=0,\Yt^*_{i_1:i_2}=1\}}.
\end{equation*}
Write for short $\widetilde{d}(\Yt)=\sum_{\substack{1\leq i_1<i_2\leq n\\}}\mathbb{I}_{\{\Yt_{i_1:i_2}=0,\Yt^*_{i_1:i_2}=1\}}$. Here $\widetilde{d}(\Yt)$ can be understood as a version of $\dY$ when we restrict $m=2$. By the above inequality,  we have the following relationship between $\dY$ and $\widetilde{d}(\Yt)$
\begin{equation}\label{UBpf3:1}
   \widetilde{d}(\Yt)\leq \tbinom{k}{2}\dY/\tbinom{k}{m},\qquad \mbox{ for all } 2\leq m \leq k.
\end{equation}
Now we are ready to derive a bound for the number of hypotheses $y$ that satisfy $\dY=t$ by the following lemma
\begin{lemma}\label{lemma:UB2} For each $\widetilde{t}$, we have
\begin{equation*}
   \big|\big\{y\in Y:\widetilde{d}(\Yt)\leq \widetilde{t}\big\}\big|\leq n^{16\widetilde{t}/k}.
\end{equation*}
\end{lemma}
Combining this with (\ref{UBpf3:1}) and elementary algebra 
\begin{equation*}
   \big|\big\{y\in Y:\dY=t\big\}\big|\leq \big|\big\{y\in Y:\dY\leq t\big\}\big|\leq \big|\big\{y\in Y:\widetilde{d}(\Yt)\leq \tbinom{k}{2}t/\tbinom{k}{m}\big\}\big|\leq n^{8kt/\tbinom{k}{m}}.
\end{equation*}
This finishes the proof.

\subsection{Proof of Lemma \ref{lemma:UB2}}
We follow a similar proof as in \cite{chen2014statistical} by first deriving a bound for the number of non-isolated nodes.

Denote the ground truth community partition associated with $y^*$ by $(C^*_1,\ldots, C^*_{r+1})$, where $C^*_i$ is the set of the nodes in the community $i$, $1\leq i \leq r$ and $C^*_{r+1}$ is the set of isolated nodes. Note that the partition hypotheses are equivalent if they are only different in the order of partitioned communities. Without loss of generality, we can reorder the partition $(C_1,\ldots, C_{r+1})$ associated with $y$ as $(\widetilde{C}_1,\ldots,\widetilde{C}_{r+1})$ in the following way:
\begin{itemize}
    \item Let $\widetilde{C}_{r+1}=C_{r+1}$ be the set of isolated nodes in the membership partition hypothesis $y$.
    \item For the $r$ community partition sets $\widetilde{C}_1,\ldots,\widetilde{C}_{r}$, 
    \begin{itemize}
        \item (a) if there exists  $i \in \{1,\ldots,r\}$ such that $|C_j\cap C^*_i|> \frac{k}{2}$, we let $\widetilde{C}_i=C_j$, $1\leq j \leq r$; 
        \item (b) the remaining sets not processed in (a) are assigned arbitrarily such that $(\widetilde{C}_1,\ldots,\widetilde{C}_{r+1})$ is a partition of ${\cal V}$.
    \end{itemize}
\end{itemize}
Note that the way of labelling in (a) is unique since the size of each community is $k$. It can be observed that this new partition $(\widetilde{C}_1,\ldots,\widetilde{C}_{r+1})$ satisfies the following properties:
\begin{itemize}
    \item  The sets in $(\widetilde{C}_1,\ldots,\widetilde{C}_{r+1})$ are mutually disjoint and $|\widetilde{C}_i|=k$  for all $1\leq i \leq r$.
    \item For each $i\in \{1,\ldots,r\}$, either $|\widetilde{C}_i\cap C_i^*|>\frac{k}{2}$, or $|\widetilde{C}_j\cap C^*_i|\leq \frac{k}{2}$ for all $j\in \{1,\ldots,r\}$.
\end{itemize}
Recall that $\widetilde{d}(\Yt)$ counts all the node pairs $(i_1,i_2)$ such that $\Yt_{i_1:i_2}^*=1$ but $\Yt_{i_1:i_2}=0$. By our definitions and elementary combinatorial calculations
\begin{equation*}
    \widetilde{d}(\Yt)=\sum_{1\leq i \leq  r}(N_1(i)+N_2(i)),
\end{equation*}
where 
\begin{align*}
    N_1(i)= M_i \cdot |C_i^*\cap \widetilde{C}_{i}|,\qquad N_2(i)= \tbinom{M_i}{2}-\sum_{\substack{j\in[r]\backslash \{i\}}}\tbinom{|C_i^*\cap \widetilde{C}_{j}|}{2}, \qquad \mbox{with } M_i=\sum_{j\in[r+1]\backslash\{i\}}|C_i^*\cap \widetilde{C}_{j}|.
\end{align*}
Here $M_i$ counts the number of misclassified nodes of community $i$ in hypothesis $y$; $N_1(i)$ counts the node pairs $(k_1,k_2)$ where $k_1,k_2$ are in the true partition set $C^*_i$ but one of them is assigned to a partition set in $(\widetilde{C}_1,\ldots,\widetilde{C}_{r+1})\backslash \widetilde{C}_i$ and the rest $m-1$ nodes stay in $\widetilde{C}_i$ in partition hypothesis $y$; $N_2(i)$ counts the node pairs $(k_1,k_2)$ that are in the the true partition set $C^*_i$ but all of them are assigned to the  isolated-node partition set $\widetilde{C}_{r+1}$ or  two different partition sets in $(\widetilde{C}_1,\ldots,\widetilde{C}_{r+1})\backslash \widetilde{C}_i$.

By the condition of this lemma, $ \widetilde{d}(\Yt)\leq \widetilde{t}$. It follows that
\begin{equation}\label{UBpf4:1}
    \sum_{i=1}^r\big(N_1(i)+N_2(i)\big) \leq \widetilde{t}.
\end{equation}
We are going to derive a bound on $\sum_{i=1}^r M_i$ based on the above properties. For each $i$, consider the cases $|\widetilde{C}_i\cap C_i^*| > \frac{1}{4} k$ and $|\widetilde{C}_i\cap C_i^*| \leq \frac{1}{4} k$ separately.

On one hand, if $|C_i\cap C_i^*| > \frac{1}{4} k$,  we have
\begin{equation}\label{ak}
     N_1(i)=M_i \cdot |C_i^*\cap \widetilde{C}_{i}|>\frac{1}{4}kM_i.
\end{equation}
On the other hand, if $|\widetilde{C}_i\cap C_i^*| \leq \frac{1}{4} k \leq \frac{1}{2}k$, by the property of $\widetilde{C}_i$, it must have $|\widetilde{C}_j\cap C_i^*| \leq  \frac{1}{2}k$ for all $1 \leq j \leq r$ and therefore $M_i=k-|\widetilde{C}_i\cap C_i^*|>\frac{3}{4}k$. It follows that
    \begin{align*}
        \begin{split} \label{1-ak}
         N_2(i) = &\tbinom{M_i}{2}-\sum_{\substack{j\in[r]\backslash \{i\}}}\tbinom{|C_i^*\cap \widetilde{C}_{j}|}{2}\\
         =&\frac{M_i^2}{2}-\sum_{\substack{j\in[r]\backslash \{i\}}}\frac{|C_i^*\cap \widetilde{C}_{j}|^2}{2}-\frac{|C_i^*\cap \widetilde{C}_{r+1}|}{2}\\
            \geq&  \frac{M_i^2}{2}-\frac{k}{2}\sum_{j\in[r+1]\backslash\{i\}}|C_i^*\cap \widetilde{C}_{j}|\\
            =& \frac{M_i^2}{2}-\frac{k}{2}M_i,
        \end{split}
    \end{align*}
where the RHS is larger than $\frac{1}{4}kM_i$ by $M_i>\frac{3}{4}k$. Combining this with (\ref{ak}), it can be observed that we always have $N_1(i)+N_2(i)\geq \frac{1}{4}kM_i$ for $1\leq i \leq r$ and therefore by (\ref{UBpf4:1})
\begin{equation*}
    \widetilde{t} \geq \frac{k}{4}\sum_{1\leq i\leq r}M_i.
\end{equation*}
Write for short $N=4\widetilde{t}/k$. Thus, we have $\sum_{i=1}^r M_i \leq N$, i.e., the total number of misclassified non-isolated nodes is upper bounded by $ N$. This means that the total number of misclassified isolated nodes is upper bounded by $N$ since each misclassified isolated node must result in a misclassified non-isolated node. Based this resulting constraint, we can bound the size of the hypotheses set $\big\{y\in Y:\widetilde{d}(\Yt)\leq\widetilde{t}\big\}$. %(\ref{CB}).

For a fixed number of misclassified non-isolated nodes ( denoted by $n_1$) and isolated nodes ( denoted by $n_2$), there are no more than $\tbinom{rk}{n_1}\tbinom{n-rk}{n_2}$  different ways to choose these misclassified nodes. At the same time,  each misclassified  node is reassigned to one of $r$ other partition sets and thus there are no more than $r^{n_1+n_2}$ cases. Therefore,
\begin{equation*}
    \big|\big\{y\in Y:\widetilde{d}(\Yt)\leq\widetilde{t}\big\}\big|\leq \sum_{n_1,n_2}\tbinom{rk}{n_1}\tbinom{n-rk}{n_2}r^{n_1+n_2}.
\end{equation*}
Note that $n_1,n_2\leq N$.  It follows that
\begin{equation*}
    D_t\leq \sum_{n_1,n_2}\tbinom{rk}{n_1}\tbinom{n-rk}{n_2}r^{n_1+n_2}\leq \sum_{n_1,n_2}\tbinom{rk}{n_1}\tbinom{n-rk}{n_2}n^{2n}\leq  n^{2N}n^{2N}=n^{16\widetilde{t}/k},
\end{equation*}
which holds for each $\widetilde{t}$. This proves Lemma \ref{lemma:UB1}.

%%%%%%%%%%%%%%%%%%%%%%%%%
%%%%%%%%%%%%%%%%%%%%%%%%%
%%%%%%%%%%%%%%%%%%%%%%%%%
\subsection{Proof of Theorem \ref{thm:UB2}}
We only prove the case when $q<p$, the proof for $p<q$ is similar and therefore we omit the proof.

Using the same notation as that in the proof of Theorem \ref{thm:UB}, similarly, to prove the claim of this theorem, it is sufficient to show that
\begin{equation}\label{UB2pf1}
    \sum_{ \tbinom{k-1}{m-1}\leq t \leq  r\tbinom{k}{m}}\mathbb{P}(E_t|y^*)=o_n(1),\qquad\mbox{with}\qquad E_t=\big\{\exists\, y\in \{y\in Y: d(\Yt)=t\},(I)\geq (p-q)\dY\big\}
\end{equation}
where
\begin{equation*}
       d(\Yt)=\sum_{\substack{1\leq i_1<\cdots<i_m\leq n}}\mathbb{I}_{\{\Yti^*\neq \Yti\}}/2,\qquad\mbox{with}\quad(I)=\sum_{\substack{1\leq i_1<\cdots<i_m\leq n}}(\Ai-\mathbb{E}[\Ai])(\Yti-\Yti^*).
\end{equation*}
Fixing $t$, we are going to derive a bound for $\mathbb{P}(E_t|y^*)$. 

First we evaluate the probability of $(I)\geq (p-q)\dY$ given hypothesis $y$. Note that $(I)$ is a sum of $2\dY$ independent random variables with sub-Gaussian norm smaller than $\max\{\sigma_p^2,\sigma_q^2\}$. By Hoeffding's inequality, for any $t>0$
\[
\mathbb{P}((I)\geq  t|y^*,y) \leq 2e^{-\frac{2t^2}{2\dY\max\{\sigma_p^2,\sigma_q^2\}}}.
\]
Letting $t=(p-q)\dY$, it follows that
\begin{equation}\label{UB2pf2}
    \mathbb{P}((I)\geq  (p-q)\dY|y^*,y) \leq 2e^{-(p-q)^2\dY/\max\{\sigma_p^2,\sigma_q^2\}}.
\end{equation}
Now let $D_t=|\{y\in Y: d(\Yt)=t\}|$ be size of the set $\{y\in Y: d(\Yt)=t\}$. By the union bound and (\ref{UB2pf2}), we have
\begin{equation}\label{UB2pf3}
    \mathbb{P}(E_t|y^*)\leq D_t\cdot 2e^{-(p-q)^2\dY/\max\{\sigma_p^2,\sigma_q^2\}}.
\end{equation}

Now we are ready to show (\ref{UB2pf1}). Combining (\ref{UB2pf3}) with Lemma \ref{lemma:UB1} and our condition $  \frac{(p-q)^2}{\max\{\sigma_p^2,\sigma_q^2\}}\geq Ck\frac{\log(n)}{\tbinom{k}{m}}$, where $C$ is  a constant, we have
\begin{align*}
    \sum_{ \tbinom{k-1}{m-1}\leq t \leq  r\tbinom{k}{m}}\mathbb{P}(E_t|y^*)\leq &   \sum_{ \tbinom{k-1}{m-1}\leq t \leq  r\tbinom{k}{m}} D_t\cdot 2e^{-(p-q)^2t/\max\{\sigma_p^2,\sigma_q^2\}}\\
    \leq &
    \sum_{ \tbinom{k-1}{m-1}\leq t \leq  r\tbinom{k}{m}} n^{8kt/\tbinom{k}{m}} 2e^{-C\frac{k\log(n)}{\tbinom{k}{m}}t}\\
    \leq  &\sum_{ \tbinom{k-1}{m-1}\leq t \leq  r\tbinom{k}{m}}2n^{-Ckt/\tbinom{k}{m}}\\
    \leq &Cn^{-Ck\tbinom{k-1}{m-1}/\tbinom{k}{m}},
\end{align*}
where we note that the RHS is $\leq Cn^{-Cm}=o_n(1)$, which finishes the proof.

\end{document}